\pdfoutput=1

\documentclass[11pt]{article}
\usepackage[final]{style/acl}
\usepackage{setspace} 

\usepackage{times}
\usepackage{latexsym}
\usepackage[T1]{fontenc}
\usepackage[utf8]{inputenc}
\usepackage{microtype}
\usepackage{inconsolata}
\usepackage{graphicx}
\usepackage{spverbatim}
\usepackage{algorithm}
\usepackage{amsmath}
\usepackage{algpseudocode}
\usepackage{multirow}
\usepackage{float}
\usepackage{booktabs}
\usepackage{listings}[language=Python]

\title{FactGenius: Combining Zero-Shot Prompting and Fuzzy Relation Mining \\to Improve Fact Verification with Knowledge Graphs
}

\author{
  Sushant Gautam\\
   Simula Metropolitan Center for Digital Engineering\\ 
   Oslo, Norway \\
  \texttt{sushant@simula.no}
}

\begin{document}
\maketitle

\hyphenpenalty=10000

\begin{abstract}
Fact-checking is a crucial natural language processing (NLP) task that verifies the truthfulness of claims by considering reliable evidence. Traditional methods are often limited by labour-intensive data curation and rule-based approaches. In this paper, we present FactGenius, a novel method that enhances fact-checking by combining zero-shot prompting of large language models (LLMs) with fuzzy text matching on knowledge graphs (KGs). Leveraging DBpedia, a structured linked data dataset derived from Wikipedia, FactGenius refines LLM-generated connections using similarity measures to ensure accuracy. The evaluation of FactGenius on the FactKG,  a benchmark dataset for fact verification, demonstrates that it significantly outperforms existing baselines, particularly when fine-tuning RoBERTa as a classifier. The two-stage approach of filtering and validating connections proves crucial, achieving superior performance across various reasoning types and establishing FactGenius as a promising tool for robust fact-checking.
The code and materials are available at \href{https://github.com/SushantGautam/FactGenius}{https://github.com/SushantGautam/FactGenius}.

\end{abstract}

\section{Introduction}

\begin{figure*}[!h]
    \centering
\includegraphics[width=0.9\textwidth]{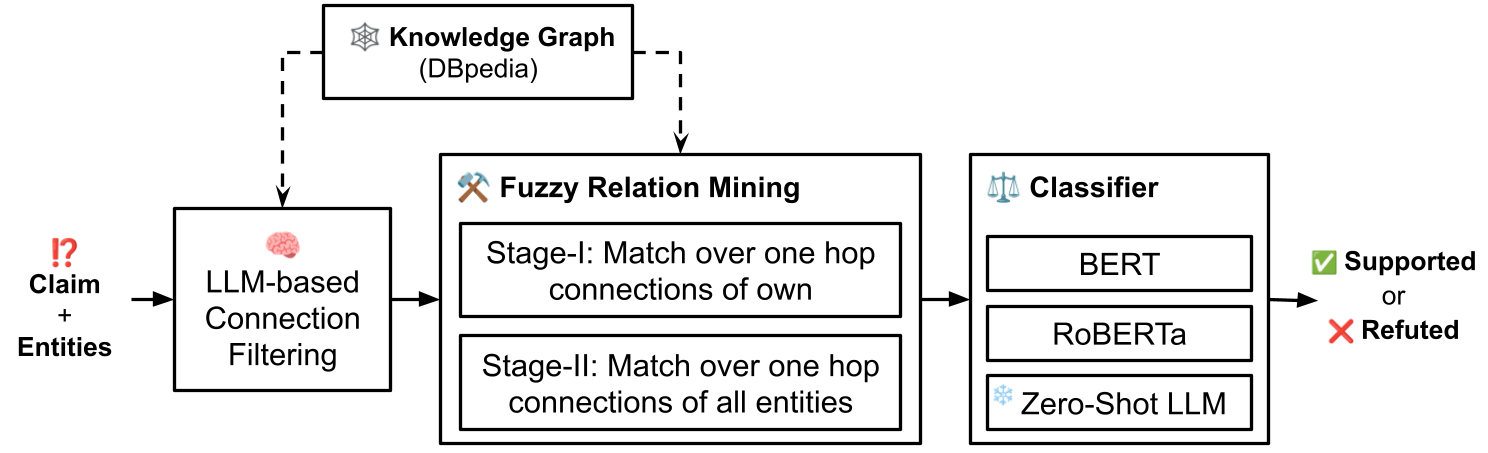}
    \caption{Overall pipeline of FactGenius: The process starts with LLM-based Connection Filtering using a knowledge graph (see Section \ref{Sec:FilteringPossibleConnections}). In Fuzzy Relation Mining (see Section \ref{Sec:FuzzyRelationMining}), Stage-I matches one-hop connections of entities, and optionally, Stage-II includes all entities' connections. The classifier (BERT, RoBERTa, or Zero-Shot LLM; see Section \ref{sec:With-Evidence-Classifier}) then determines if the claim is supported or refuted.}
    \label{fig:pipeline}
\end{figure*}
Fact-checking is a critical task in natural language processing (NLP) that involves automatically verifying the truthfulness of a claim by considering evidence from reliable sources~\cite{Thorne2018Jun}. This task is essential for combating misinformation and ensuring the integrity of information in digital communication~\cite{Cotter2022Jan}. Traditional fact-checking methods rely heavily on manually curated datasets and rule-based approaches, which can be labour-intensive and limited in scope~\cite{Papadopoulos2024Mar}.

Recent advancements in large language models (LLMs) have shown promise in enhancing fact-checking capabilities~\cite{Choi2024May}. LLMs, with their extensive pre-training on diverse textual data, possess a vast amount of embedded knowledge~\cite{Yang2024Apr}. However, their outputs can sometimes be erroneous or lacking in specificity, especially when dealing with complex reasoning patterns required for fact-checking. External knowledge, such as knowledge graphs (KGs)~\cite{Hogan2021Jul}, can aid in fact-checking.

In this paper, we propose FactGenius, a novel approach that combines zero-shot prompting of LLMs with fuzzy relation-mining technique to improve reasoning on knowledge graphs. Specifically, we leverage DBpedia~\cite{Lehmann2015Jan}, a structured source of linked data, to enhance the accuracy of fact-checking tasks. 

Our methodology involves using the LLM to filter potential connections between entities in the KG, followed by refining these connections through Levenshtein distance-based fuzzy matching. This two-stage approach ensures that only valid and relevant connections are considered, thereby improving the accuracy of fact-checking.

We evaluate our method using the FactKG dataset~\cite{Kim2023Jul}, which comprises 108,000 claims constructed through various reasoning patterns applied to facts from DBpedia. Our experiments demonstrate that FactGenius significantly outperforms existing baselines~\cite{Kim2023Dec}, particularly when fine-tuning RoBERTa~\cite{Liu2019Jul} as a classifier, achieving superior performance across different reasoning types.

In summary, the integration of LLMs with KGs and the application of fuzzy matching techniques represent a promising direction for advancing fact-checking methodologies. Our work contributes to this growing body of research by proposing a novel approach that effectively combines these elements, yielding significant improvements in fact-checking performance.

\setstretch{1.05}
\newpage
\section{Literature Review}
 
Fact-checking has become an increasingly vital aspect of natural language processing (NLP) due to the proliferation of misinformation in digital communication~\cite{Guo2022Dec}. Traditional approaches to fact-checking have typically relied on manually curated datasets and rule-based methods, which, while effective in controlled environments, often struggle with scalability and adaptability to new types of misinformation~\cite{Saquete2020Mar, Guo2022Dec}. The labour-intensive nature of these methods also poses significant challenges in rapidly evolving information landscapes~\cite{Nakov2021, Zeng2021Oct}.

To address challenges in understanding machine-readable concepts in text, FactKG introduces a new dataset for fact verification with claims, leveraging knowledge graphs, encompassing diverse reasoning types and linguistic patterns, aiming to enhance reliability and practicality in KG-based fact verification~\cite{Kim2023Jul}.
Similarly, the Fact Extraction and VERification (FEVER) dataset~\cite{Thorne2018Jun} pairs claim with Wikipedia sentences that support or refute them, providing a benchmark for fact-checking models. The authors employed a combination of natural language inference models and information retrieval systems to assess claim veracity.
The GEAR framework~\cite{Zhou2019Jul} improves fact verification by using a graph-based method to aggregate and reason over multiple pieces of evidence, surpassing previous methods by enabling evidence to interact.

Recent advancements in large language models (LLMs) have demonstrated considerable potential in enhancing fact-checking processes~\cite{Kim2023Dec, Choi2024May}. LLMs have been pre-trained on vast and diverse corpora~\cite{Yang2024Apr}, enabling them to generate human-like text and possess a broad knowledge base~\cite{Choi2024May}. However, despite their impressive capabilities, LLMs can produce outputs that are erroneous or lack the specificity required for complex fact-checking tasks~\cite{Choi2024May}. This is particularly evident when intricate reasoning and contextual understanding are necessary to verify claims accurately~\cite{Chai2023Oct}.
Several studies have explored the integration of LLMs with external knowledge sources to improve their performance in fact-checking tasks~\cite{Cui2023Jun, Ding2023Dec}.

The incorporation of knowledge graphs (KGs) into fact-checking frameworks has also garnered attention. KGs, such as DBpedia~\cite{Lehmann2015Jan}, provide structured and linked data that can enhance the contextual understanding of LLMs. 

Knowledge graphs have been used to improve various NLP tasks by providing additional context and relationships between entities, as demonstrated by initiatives for knowledge-aware language models ~\cite{Li2023Oct, LoganIv2019Jun} and KG-BERT ~\cite{Yao2019Sep}.

Approximate string matching~\cite{Navarro2001Mar}, also called fuzzy string matching, is a technique used to identify partial matches between text strings. Fuzzy matching techniques~\cite{Navarro2001Mar} have been applied to enhance the integration of LLMs and KGs~\cite{Wang2024Mar}.

\setstretch{1.0}
Levenshtein distance-based similarity measure~\cite{levenshtein1966binary} helps in identifying strings which have approximate matches which can be useful to find relevant connections between entities by accommodating minor discrepancies in data representation
This approach has been beneficial in refining the outputs of LLMs, ensuring that only valid and contextually appropriate connections are considered~\cite{Guo2023Oct}.

Our proposed method, FactGenius, builds on these advancements by combining zero-shot prompting of LLMs with a fuzzy relation-mining technique to improve reasoning over KGs. This methodology leverages DBpedia as a structured source of linked data to enhance fact-checking accuracy. By using LLMs to filter potential connections between entities and refining these connections through fuzzy matching, FactGenius aims to address the limitations of existing fact-checking models.

\setstretch{1.0}
\section{Methodology}
FactGenius leverages the capabilities of a Large Language Model (LLM) to filter possible connections between entities in a Knowledge Graph (KG), particularly utilizing DBpedia~\cite{Lehmann2015Jan} as a structured source of linked data.

Since the output of LLMs can be erroneous, the connections are further refined and enriched using Levenshtein distance~\cite{levenshtein1966binary} and are also validated to ensure that such connections exist.
This process is crucial for tasks such as fact-checking, where establishing valid and relevant connections between entities can validate or refute claims. Finally, the classifier, which can be fine-tuned over pre-trained models like BERT~\cite{Devlin2019Jun} or RoBERTa~\cite{Liu2019Jul}, or a Zero-Shot LLM, determines whether the claim is supported or refuted. The overall pipeline is shown in  Figure \ref{fig:pipeline}.

\subsection{Dataset}
The FactKG dataset~\cite{Kim2023Jul} is used which comprises 108,000 claims constructed through various reasoning patterns applied to facts sourced from DBpedia~\cite{Lehmann2015Jan}. 
Each data point consists of a natural language claim in English, the set of DBpedia entities mentioned in the claim, and a binary label indicating the claim's veracity (Supported or Refuted). The distribution across labels and five different reasoning types is shown in Table \ref{tab:data-distribution}. The relevant relation paths starting from each entity in the claim are known which aids in the evaluation and development of models for claim verification tasks. 

The dataset is accompanied by processed DBpedia, an undirected knowledge graph (KG). The dataset provides researchers with a valuable resource for exploring the intersection of natural language understanding and knowledge graph reasoning.

\begin{table}[htbp]
\centering
\caption{Data distribution across labels and five reasoning types. }
\label{tab:data-distribution}
\begin{tabular}{|l|c|c|c|}
\hline
 Set& \textbf{Train} & \textbf{Valid} & \textbf{Test} \\ \hline
\textbf{Total Rows} & 86367 & 13266 & 9041 \\ \hline
True  (Supported) & 42723 & 6426 & 4398 \\ \hline
False (Refuted) & 43644 & 6840 & 4643 \\ \hline \hline
\textbf{One-hop} & 15069 & 2547 & 1914 \\ \hline
\textbf{Conjunction} & 29711 & 4317 & 3069 \\ \hline
\textbf{Existence} & 7372 & 930 & 870 \\ \hline
\textbf{Multi Hop} & 21833 & 3555 & 1874 \\ \hline
\textbf{Negation} & 12382 & 1917 & 1314 \\ \hline
\end{tabular}
\end{table}

\subsection{Claim Only Classifier}
In this setting, where the models are given only the claim and tasked with predicting the label, it is expected that the model will heavily depend on stored evidence within its trained weights or identify patterns within the structure of the provided claims.

\subsubsection{Zero-shot Claim Only Baseline}
 A baseline is established using the Meta-Llama-3-8B-Instruct\footnote{\href{https://huggingface.co/meta-llama/Meta-Llama-3-8B-Instruct}{huggingface.co/meta-llama/Meta-Llama-3-8B-Instruct}}~\cite{Llama3} model with zero-shot promoting for claim verification, asking it to verify the claim without evidence. Through instruction prompt engineering, it is ensured that the model responds with either 'true' or 'false'. A retry mechanism is implemented to handle potential failures in LLM responses. A prompt example is shown in Figure \ref{llm-prompt-zero-shot-claim-only}.

 \begin{figure}[htpb]
\fontsize{8}{6}\selectfont
\begin{spverbatim}
[{
"role":"system", "content": 
"You are an intelligent fact checker trained on Wikipedia. You are given a single claim and your task is to decide whether all the facts in the given claim are supported by the given evidence using your knowledge.
Choose one of {True, False}, and output the one-sentence explanation for the choice. "
},{
"role":"user", "content":
'''
## TASK:
Now let’s verify the Claim based on the evidence.
Claim:  
< < < Well, The celestial body known as 1097 Vicia has a
mass of 4.1kg.> > > 

#Answer Template: 
"True/False (single word answer),
One-sentence evidence."
'''
}]


\end{spverbatim}
\caption{Example prompt given to Llama3-Instruct without evidence for zero-shot fact-checking. \\ \small < < < ... > > > signs are added just to indicate that the content inside is different for each prompt. }
\label{llm-prompt-zero-shot-claim-only}
\end{figure}

\subsubsection{RoBERTa as Claim Only Fact Classifier}
RoBERTa-base\footnote{\href{https://huggingface.co/FacebookAI/roberta-base}{huggingface.co/FacebookAI/roberta-base}} is fine-tuned with claims as input, training it to predict Supported or Refuted. This is to compare with the BERT baseline reported in previous works~\cite{Kim2023Jul}.

\begin{figure}[h!]
\fontsize{8}{6}\selectfont
\begin{spverbatim}
[{
"role":"system", "content": 
"You are an intelligent graph connection finder. You are given a single claim and connection options for the entities present in the claim. Your task is to filter the Connections options that could be relevant to connect given entities to fact-check Claim1. ~ ( tilde ) in the beginning means the reverse connection. "
},{
"role":"user", "content":
'''
Claim1:
<<<Well, The celestial body known as 1097 Vicia has a mass of 4.1kg.>>>
      
## TASK:
- For each of the given entities given in the DICT structure below: 
    Filter the connections strictly from the given options that would be relevant to connect given entities to fact-check Claim1.
- Think clever, there could be multi-step hidden connections, if not direct, that could connect the entities somehow.
- Prioritize connections among entities and arrange them based on their relevance. Be extra careful with ~ signs.
- No code output. No explanation. Output only valid python DICT of structure:

{
<<<
"1097_Vicia": ["...", "...", ... ],  
# options (strictly choose from): discovered, formerName, epoch, periapsis, apoapsis, ..., Planet/temperature

"4.1": ["...", "...", ... ],  
# options (strictly choose from): ~length, ~ethnicGroups, ~percentageOfAreaWater, ~populationDensity, ~engine, ..., ~number
}
>>>
'''
}]


\end{spverbatim}
\caption{Example prompt given to Llama3-Instruct to filter potential connections between entities based on a given claim.}
\label{llm-prompt-filter}
\end{figure}

\subsection{Graph Filtering}
\label{sec:Graph-Filtering}
The graph filtering is divided into two main stages:

\subsubsection{Filtering Possible Connections}
\label{Sec:FilteringPossibleConnections}
This stage involves utilizing an LLM, particularly the Llama3-Instruct model, to identify and filter potential connections between entities based on a given claim. The detailed steps are as follows:

\textbf{Data Preparation}
Entity sets and their possible connections are extracted from the KG (DBpedia). Each entity and its associated possible connections form the initial input for the LLM.


\textbf{LLM Integration}
The LLM is tasked with identifying relevant connections for each entity in the specific claim. The process involves:
\begin{enumerate}
    \item Encoding each entity and its possible connections into a structured format suitable for the LLM.
    \item Utilizing the LLM's inference capabilities to filter out irrelevant connections based on the context provided by the claim.
    \item Generating a filtered set of connections in a structured format, which is then evaluated for completeness and relevance.
\end{enumerate}

An example of the prompt used with LLM in Stage-I is shown in Figure \ref{llm-prompt-filter}.
Prompts are crafted through iterative testing and refinement, aiming to optimize results and performance.

\textbf{Handling Invalid LLM Response}
A retry mechanism is implemented to handle potential failures in LLM responses. If the LLM output is inadequate (e.g., empty or nonsensical), the request is retried up to a specified maximum number of attempts, typically 10.
Throughout this experiment, however, we did not encounter any cases where the retry exceeded this limit.
 
\subsubsection{Fuzzy Relation Mining}
\label{Sec:FuzzyRelationMining}
The LLM-filtered connections are then validated against the KG to ensure their existence and relevance. This involves:
\begin{enumerate}
    \item \textbf{Stage-I}:  Checking each connection filtered using LLM against the KG to confirm its validity. For each connection in the entities, perform fuzzy matching using Levenshtein distance to match entities in the first-hop relation of the graph. This approach accommodates speckling and reverse connection errors.
    \item \textbf{Stage-II}: Matching potential connections fuzzily, while considering reverse relationships and similarities across all the one-hop connections in the knowledge graph of all entities within the claims.
\end{enumerate}
The details are explained in Algorithm \ref{alg:validateRelation}.

\begin{algorithm}[!h]
\caption{Relationship Mining with Validation}
\fontsize{8}{6}\selectfont
\label{alg:validateRelation}
\begin{algorithmic}[1]
\State \textbf{Input:} $A$ - dictionary of entities with their connections, $G$: Graph
\State \textbf{Output:} $probable\_connections$- dictionary of entities with  updated and validated connections

\vspace{2mm}
\Procedure{ValidateRelation}{$A$}
    \State Initialize: $\textbf{probable\_connections}$: \{\}
    \vspace{2mm}
    \State \textbf{--- \underline{Stage-I} ---}
    \vspace{2mm}
    \For{\textbf{each} entity, connections in $A$}
        \State Retrieve: all $\textbf{one\_hop\_connections}$ for \textbf{entity} $G$
        \For{\textbf{each} connection in \textbf{connections}}
            \State Fuzzily match from $\textbf{one\_hop\_connections}$
            \State Filter matches with a similarity score greater than 90
            \State Update \textbf{entity} in $\textbf{probable\_connections}$
        \EndFor
    \EndFor
    \vspace{2mm}
    \State \textbf{--- \underline{Stage-II} (optional) ---}
    \vspace{2mm}
    \State $\textbf{all\_connections}$ = all connections in $\textbf{probable\_connections}$
    \For{\textbf{each} entity, connections in $\textbf{probable\_connections}$}
        \State Retrieve: all $\textbf{one\_hop\_connections}$ for \textbf{entity} $G$
        \For{\textbf{each} connection in $\textbf{all\_connections}$}
            \State  Fuzzily match from $\textbf {one\_hop\_connections}$
            \State Filter matches with a similarity score greater than 90
            \State Update \textbf{entity} in $\textbf{probable\_connections}$  \EndFor
    \EndFor
    \vspace{2mm}
\EndProcedure
\end{algorithmic}

\end{algorithm}

\setstretch{1.15}

\subsection{With Evidence Classifier}
\label{sec:With-Evidence-Classifier}
In this configuration, the model is supplied with both the claim and graphical evidence as input, and it then makes predictions regarding the label. 
FactGenius utilizes graph filtering, as explained in Section \ref{sec:Graph-Filtering}, to ensure retention of the most relevant and accurate connections.

\subsection{Zero-shot LLM as Fact Classifier}
This involves utilizing Llama-3-Instruct as a fact classifier, to predict Supported or Refuted for the given input claim and evidence.  A retry mechanism is implemented to handle potential failures in LLM responses. A prompt example with evidence is shown in Figure \ref{llm-prompt-zero-shot-with-evidence}.

\subsection{Fine-tuning pre-trained models}
Pre-trained BERT-base-uncased\footnote{\href{https://huggingface.co/google-bert/bert-base-uncased}{huggingface.co/google-bert/bert-base-uncased}} and RoBERTa-base are finetuned with claim and evidence as inputs to predict whether the claim is supported or refuted.
In addition, an ablation evaluates the contributions of each stage of our approach. This involved sequentially removing Stage-II and measuring the performance of the system after the removal. The results of the ablation study allowed us to quantify the impact of both stages on the overall performance of the model. Accuracy as an evaluation metric across all reasoning types was employed to quantify the performance improvements resulting from the ablation study.

\begin{figure}[htpb]
\fontsize{8}{6}\selectfont
\begin{spverbatim}
[{
"role":"system", "content": 
"You are an intelligent fact-checker. You are given a single claim and supporting evidence for the entities present in the claim, extracted from a knowledge graph. 
Your task is to decide whether all the facts in the given claim are supported by the given evidence.
Choose one of {True, False}, and output the one-sentence explanation for the choice. "
},{
"role":"user", "content":
'''
## TASK:
Now let’s verify the Claim based on the evidence.
Claim:  
< < < Well, The celestial body known as 1097 Vicia has a
mass of 4.1kg.> > > 

Evidences: 
< < <
1999_Hirayama >- mass -> ""4.1""
1097_Vicia >- mass -> ""9.8"""
> > > 

#Answer Template: 
"True/False (single word answer),
One-sentence evidence."
'''
}]


\end{spverbatim}
\caption{Example prompt given to Llama3-Instruct with evidence for zero-shot fact-checking. }
\label{llm-prompt-zero-shot-with-evidence}
\end{figure}

\subsection{Implementation}
Our FactGenius system implementation leverages several advanced tools and frameworks to ensure efficient and scalable processing. The Llama3-Instruct inference server is set up using vLLM~\cite{vllm,vllmpaper}, running on an NVIDIA A100 GPU (80 GB vRAM) to facilitate rapid inference. This server runs standalone, integrating seamlessly with the FactGenius pipeline.

For model fine-tuning and evaluation, we employ the Hugging Face Transformers library, utilizing the \texttt{Trainer} class for managing the training process. This setup allows for the fine-tuning of pre-trained models like BERT and RoBERTa on our pipeline. Hyper-parameters such as batch size, learning rate, and training epochs are configured to optimize performance, with computations accelerated by PyTorch.

\setstretch{1.0}

\section{Results}
\setstretch{0.99}
To evaluate the performance of our proposed methods, we conducted a series of experiments comparing different strategies for fact-checking. The results are summarized in Table \ref{tab:results}.
\begin{table*}[t]
\centering
\caption{Comparing our method with other strategies and methods in terms of reported accuracies in the test set. \\
\small{* indicates results obtained from KG-GPT paper~\cite{Kim2023Dec}.}
}
\label{tab:results}
\resizebox{\textwidth}{!}{%
\begin{tabular}{@{}|c|c|c|c|c|c|c|c|c|@{}}
\toprule
Input type & Model & Variants & One-hop & Conjunction & Existence & Multi-hop & Negation & Total \\ \midrule
\multirow{4}{*}{Claim Only} & baseline & Llama3-Instruct-zero-shot  & 0.61 & 0.67 & 0.59 & 0.61 & 0.53 & 0.61 \\ \cmidrule(l){2-9}
& Fact-KG & BERT* & 0.69 & 0.63 & 0.61 & 0.70 & 0.63 & 0.65 \\ \cmidrule(l){2-9}
& KG-GPT & ChatGPT (12-shot)* & - & - & - & - & - & 0.68 \\ \cmidrule(l){2-9}
& baseline & RoBERTa & 0.71 & 0.72 & 0.52 & 0.74 & 0.54 & 0.68 \\ \midrule
\multirow{6}{*}{With Evidence} & Fact-KG & GEAR* & 0.83 & 0.77 & 0.81 & 0.68 & 0.79 & 0.77 \\ \cmidrule(l){2-9}
& KG-GPT & KG-GPT (12-shot)* & - & - & - & - & - & 0.72 \\ \cmidrule(l){2-9}
& \multirow{4}{*}{FactGenius} & Llama3-Instruct-zero-shot & 0.72 & 0.75 & 0.76 & 0.62 & 0.52 & 0.68 \\ \cmidrule(l){3-9}
& & BERT-two-stage & 0.75 & 0.67 & 0.94 & 0.66 & 0.79 & 0.72 \\ \cmidrule(l){3-9}
& & RoBERTa-single-stage & 0.87 & 0.82 & 0.94 & 0.75 & 0.84 & 0.83 \\ 
\cmidrule(l){3-9}
& & \textbf{RoBERTa-two-stage} & \textbf{0.89} & \textbf{0.85} & \textbf{0.95} & \textbf{0.75} & \textbf{0.87} & \textbf{0.85} \\ 
\bottomrule
\end{tabular}%
}
\end{table*}

\begin{figure}[th]
    \centering
    \includegraphics[width=0.29\textwidth]{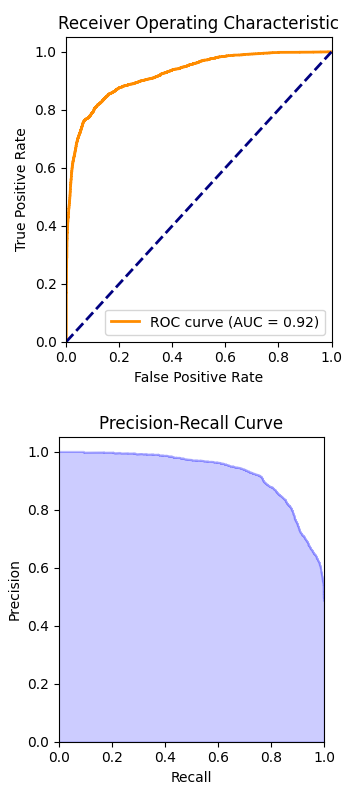}
    \caption{ROC curve (top) illustrates classifier discrimination ability with AUC 0.92, while Precision-Recall curve (bottom) reveals precision across recall levels for the best model in the test dataset.}
    \label{fig:roc-auc}
\end{figure}

\begin{table}[th]
\centering
\caption{Classification report for the best model across the test dataset.}
\label{fig:classification-report}
\begin{tabular}{p{1.1cm}cccc}
\toprule
 & \textbf{Precision} & \textbf{Recall} & \textbf{F1} & \textbf{Support} \\
\midrule
\textbf{Refuted} & 0.81 & 0.93 & 0.86 & 4643 \\
\textbf{Supported} & 0.91 & 0.77 & 0.83 & 4398 \\
\midrule
\textbf{Accuracy} & 0.86 & 0.85 & 0.85 & 9041\\
(average)  & &  & &  \\
\end{tabular}
\end{table}

\begin{table}[h!]
\centering
\caption{Confusion matrix for the best model across the test dataset.}
\label{fig:confusion-matrix}
\begin{tabular}{cc|cc}
 & & \multicolumn{2}{c}{\textbf{\underline{Predicted}}} \\
 & & \textbf{Refuted} & \textbf{Supported} \\
\hline
\multirow{2}{*}{\textbf{\underline{Actual}}} & \textbf{Refuted} & 4315 & 328 \\
 & \textbf{Supported} & 1031 & 3367 \\
\end{tabular}
\end{table}

\subsection{Baseline and Claim Only Models}
This achieved an accuracy of 0.61, demonstrating the inherent knowledge embedded within the Llama3 about the facts in our corpus. 
Adding evidence to the Llama3-Instruct model's instructions significantly improved its accuracy from 0.61 to 0.68.
This reflects the trivial phenomenon that, incorporating relevant evidence can enhance fact-checking performance in a zero-shot learning scenario where the performance is mostly dependent on knowledge embedded in the model.

\subsection{Comparison of Different Models}

We compared the performance of RoBERTa, under the claim-only scenario. RoBERTa outperformed the reported accuracy of BERT~\cite{Kim2023Jul}, achieving an accuracy of 0.68, which is on par with the 12-shot ChatGPT model reported in the KG-GPT paper \cite{Kim2023Dec}. This suggests that RoBERTa is a highly effective model for fact-checking tasks.

\subsection{Incorporating Evidence with FactGenius}

 We employed a zero-shot prompting technique to filter relevant connections from the evidence, followed by fuzzy matching across multiple stages. This approach enabled us to retrieve evidence by searching over the graph. However, directly applying zero-shot prompting with Llama3-Instruct, even with evidence, did not yield superior performance.
When using fine-tuned BERT as a classifier, the performance was comparable to the 12-shot KG-GPT model. However, fine-tuning RoBERTa led to a significant performance boost, achieving an accuracy of 0.85, the highest among all models tested, even surpassing the GEAR baseline\cite{Zhou2019Jul}, which enhances fact verification by using a graph-based approach to aggregate and reason over multiple pieces of evidence.

\subsection{Two-Stage Approach}

To assess the contribution of our two-stage approach, we first apply only the first-stage graph filtering method (Stage-I) to filter the relationships, which achieved an accuracy of 0.83. Incorporating the second stage  (Stage-II) further improved the performance to 0.85. The second stage particularly enhanced performance across all reasoning types, with notable improvements in conjunction and negation tasks.
Interestingly, for the existence reasoning type, the BERT classifier performed on par with the best models, indicating its robustness for this specific task.

\subsection{Overall Performance}

FactGenius with a fine-tuned classifier, demonstrated superior performance across all reasoning types compared to previously reported accuracies. This validates the effectiveness of our multi-stage evidence retrieval and classifier fine-tuning approach in improving fact-checking accuracy. 
Refer to Figure \ref{fig:roc-auc} for the ROC and Precision-Recall curves illustrating the classifier performance of the best FactGenius variant with two-stage filtering relationship mining and fine-tuned RoBERTa classifier and to Table \ref{fig:classification-report} and \ref{fig:confusion-matrix} for the classification report and confusion matrix, respectively.

\setstretch{0.96}

\section{Discussion}

The improved performance of FactGenius, particularly in Conjunction, Existence, and Negation reasoning, can be attributed to its innovative combination of zero-shot prompting with large language models and fuzzy text matching on knowledge graphs.

The two-stage approach, which involves an initial filtering phase followed by a validation phase, significantly enhances accuracy. However, the model shows moderate performance improvement in Multi-hop reasoning, indicating the need for more advanced techniques to handle its complexity. 

The two-stage approach of filtering and validating connections proved to be particularly effective. In the first stage, the LLM helps to narrow down potential connections based on the context provided by the claim. This initial filtering significantly reduces the search space, making the subsequent validation stage more efficient. The second stage further refines these connections through fuzzy matching, ensuring that only the most relevant and accurate connections are retained. The comparative study confirmed the importance of each stage, showing that the second stage particularly enhances performance in conjunction and negation reasoning tasks.

As having an LLM inference server is a crucial component of this framework, we employed vLLM~\cite{vllm} to streamline rapid inference with a single NVIDIA A100 GPU.  In our experiment, the LLM inference speed was around 15 queries per second, including retries in case of failure. This rate is feasible, considering that LLM inference is continually optimized with the latest technologies. Embedding LLM in a framework has proven to be a wise choice.

\section{Conclusion}
In this paper, we introduced FactGenius, a novel method that combines zero-shot prompting of large language models with fuzzy relation mining for superior reasoning on knowledge graphs. 
This approach addresses several key challenges in traditional fact-checking methods. 
First, the integration of LLMs allows for the leveraging of extensive pre-trained knowledge, which is crucial for understanding and verifying complex claims through structured data from DBpedia. Second, the use of fuzzy text matching with Levenshtein distance ensures that minor discrepancies in entity names or relationships do not hinder the relationship selection process, thus improving robustness.

Our experiments on the FactKG dataset demonstrated that FactGenius significantly outperforms traditional fact-checking methods and existing baselines, particularly when fine-tuning RoBERTa as a classifier. The two-stage approach of filtering and validating connections proved crucial for achieving high accuracy across different reasoning types. 
This underscores the potential of FactGenius to improve fact-checking accuracy without requiring complex stages and components.

The findings from this study suggest that integrating LLMs with structured knowledge graphs and fuzzy matching techniques holds great promise for advancing fact-checking capabilities. Future work could explore the application of this approach to other domains and datasets, as well as the potential for incorporating additional sources of structured data to further enhance performance.

By improving the accuracy and efficiency of fact-checking, FactGenius contributes to the broader effort of combating misinformation and ensuring the reliability of information in digital communication.

\section*{Acknowledgement}
This work was accepted and presented at the 6th IN5550 Workshop on Neural Natural Language Processing (WNNLP 2024) at the University of Oslo, Norway.
It also has benefited from the Experimental Infrastructure for Exploration of Exascale Computing (eX3) at Simula, which is financially supported by the Research Council of Norway.


\bibliography{references}




\end{document}